\documentclass[letterpaper, 10 pt, conference]{ieeeconf}  

\IEEEoverridecommandlockouts                              
\usepackage{graphicx}
\usepackage{epstopdf}
\usepackage{amsmath}
\usepackage{float}
\usepackage{subfigure}
\usepackage{hyperref}
\title{\LARGE \bf
Visual-tactile Sensing for Real-time Liquid Volume Estimation in Grasping}

\author{Fan Zhu$^{1,2}$ , Ruixing Jia$^{2}$, Lei Yang$^{2}$, Youcan Yan$^{3}$, Zheng Wang$^{4,*}$, Jia Pan$^{2}$, Wenping Wang$^{2}$%
\thanks{*Corresponding Author
        {\tt\small zheng.wang@ieee.org}}%
\thanks{$^{1}$ School of Robotics, Xi'an-Jiaotong Liverpool University}%
\thanks{$^{2}$ Department of Computer Science, The University of Hong Kong}%
\thanks{$^{3}$ Department of Biomedical Engineering, City University of Hong Kong}%
\thanks{$^{4}$ Department of Mechanical and Energy Engineering, Southern University of Science and Technology}%
}

\begin{document}

\maketitle
\thispagestyle{empty}
\pagestyle{empty}

\begin{abstract}
We propose a deep visuo-tactile model for real-time estimation of the liquid inside a deformable container in a proprioceptive way. We fuse two sensory modalities, i.e., the raw visual inputs from the RGB camera and the tactile cues from our specific tactile sensor without any extra sensor calibrations. The robotic system is well controlled and adjusted based on the estimation model in real time. The main contributions and novelties of our work are listed as follows: 1) Explore a proprioceptive way for liquid volume estimation by developing an end-to-end predictive model with multi-modal convolutional networks, which achieve a high precision with an error of $\sim2$ ml in the experimental validation. 2) Propose a multi-task learning architecture which comprehensively considers the losses from both classification and regression tasks, and comparatively evaluate the performance of each variant on the collected data and actual robotic platform. 3) Utilize the proprioceptive robotic system to accurately serve and control the requested volume of liquid, which is continuously flowing into a deformable container in real time. 4) Adaptively adjust the grasping plan to achieve more stable grasping and manipulation according to the real-time liquid volume prediction.

\end{abstract}

\section{INTRODUCTION}
Recent years have witnessed great advancements in visual techniques and novel sensory designs related to robotics. Consequently, intelligent robotic systems become increasingly common in various areas, including manufacturing processes, service industry, surgery, etc\cite{jansen2009surgical,gemici2014learning}. A safe yet stable grasping has attracted, and still does, a great deal of interest over the last few decades. It demands a robotic gripper to apply sufficient force on the object to move it and keep it from broken at the same time. Many efforts have been devoted to solid objects to explore how forces would affect their behaviour  \cite{muller2003particle,mottaghi2016happens}. Very little attention has been paid to liquid containers and the estimation of their content. It still remains an under-researched area in the robotics community.

Researchers strive to exploit the intelligent robotic systems which are capable of operating at the same level of dexterity as humans and exploit the human sensory-motor synergies \cite{wen2020force}. Human hands are good at restraining and manipulating liquids and their containers on a daily basis. We can comfortably regulate the contact force when grasping the container and have an estimation of the volume of liquid inside with the comprehensive visual and tactile sensory. Since humans can manipulate and understand the liquid inside a container, we are motivated to transfer this adaptability skill to robots. Recently some works started to address the integration of sensing capabilities in robotic grasping, such as position \cite{chin2019simple,elgeneidy2018bending,gerboni2017feedback,soter2018bodily} and force sensing \cite{homberg2019robust}. Although T.N.Le \textit{et al.} \cite{le2020safe} took both contact detection and force sensing into consideration to grasp an empty paper cup, once the cup is filled with the liquid, their solution becomes insufficient to grasp the container due to the lack of understanding of liquid inside.

In this chapter, we aim to combine the visual and tactile capabilities, which humans are born with, to estimate the volume of liquid in a deformable container in real time and subsequently achieve adaptability of grasping force based on the liquid estimation.

We propose a deep learning model to fuse visuo-tactile signals in the robotic system for real-time estimation of the liquid inside a deformable container in a proprioceptive way. We fuse two sensory modalities, i.e., the raw visual inputs from the mounted RGB camera and the tactile cues from the specific tactile sensor \cite{yan2021soft} without any extra sensor calibrations. The robotic system is well controlled and adjusted based on the estimation model in real time. The main contributions and novelties of our work are listed as follows:\\
\begin{itemize}
	\item Explore a proprioceptive way for liquid volume estimation by developing an end-to-end predictive model with multi-modal convolutional networks, which achieve a high precision with an error of $\sim2$ ml in the experimental validation.
	\item Propose a multi-task learning architecture which comprehensively considers the losses from both classification and regression tasks, and comparatively evaluate the performance of each variants on the collected data and actual robotic platform.
	\item Utilize the proprioceptive robotic system to accurately serve and control the requested volume of liquid, which is continuously flowing into a deformable container in real time.
	\item Adaptively adjust the grasping plan to achieve more stable grasping and manipulation according to the real-time liquid volume prediction.
\end{itemize}

\section{Related Work}
\textbf{Volume estimation in robotic grasping.} When handling robotic grasping of a container with liquid, it is significant to understand the amount of liquid inside for subsequent manipulations. There are some prior works related to perceiving liquids from sensory feedback \cite{griffith2012object,rankin2010daytime}. Schenck et al. \cite{schenck2017visual} proposed a method to detect water from color images of pouring. To collect the images to train the detector, they use hot water and subsequently utilize thermal images to easily detect the hot water. Brandl et al. \cite{brandi2014generalizing} propose a method to estimate the amount of liquid from motion and the 3D model of the container. Most works related to volume estimation in robotic grasping entail high-level reasoning in visual domains and the liquid volume is mainly estimated when the liquid is standing with no motion. Hence, few methods can sufficiently understand and adapt to the real-time changes based on observations in a dynamic system. In this paper, the robotic grasping system can simultaneously estimate and adapt to the current state inside the deformable container while the liquid continuously flows into it.

\noindent \textbf{Fusion of visual and tactile sensing modalities.} Various tactile sensors have been recently proposed in the literature \cite{yousef2011tactile} and they have been employed in a range of ways to aid robotic grasping by fusing with visual sensing modality. For example, M.A.Lee et al. \cite{lee2019making} applied the RGB camera and a force-torque sensor for providing visual and tactile cues to establish a multimodal representations for contact-rich tasks with self-supervised learning. D. Guo et al. \cite{guo2017robotic} proposed to extract features from visual inputs and incorporate tactile readings into the dynamic grasp detection to a complete process of the robotic grasping containing the grasp planning and grasp execution stage. R. Calandra et al. \cite{calandra2017feeling} established a visuo-tactile model to predict grasp outcome by taking advantages of the Gelsight, which is a optical tactile sensor, to represent tactile features and fuse them with the visual ones. In our paper, we incorporate an alternative representation of tactile cues which are different from the prior works by utilizing the raw magnetic flux densities from the soft tactile sensor \cite{yan2021soft} and fusing them with the visual inputs to build a multi-modal model. Moreover, we distinctively apply multi-task learning to process the visual and tactile cues to train the model for real-time liquid estimation in grasping.

\section{System Architecture}
\label{sec2:systemArchitecture}
\begin{figure}[t] \centering
	\begin{center}
		\includegraphics[width=0.48\textwidth]{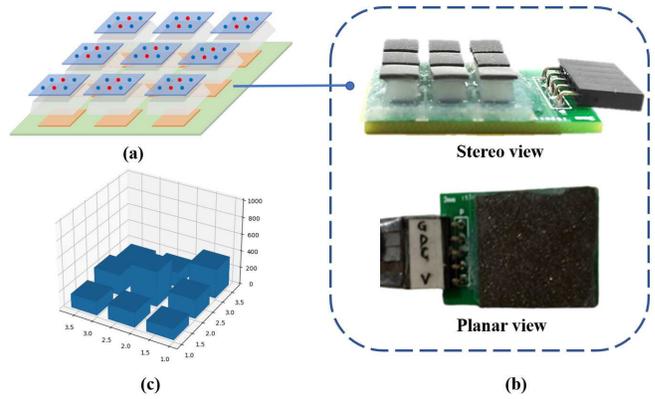}
	\end{center}
	\caption{\textbf{(a)} Illustration of the soft tactile sensor with a flat surface in a 3-by-3 array. The top layer is made of a flexible magnetic film. The middle layer is made of the soft silicone elastomer, which can sense the deformation of the magnetic film according to the change of magnetic flux densities. The bottom layer is a hard printed circuit board with a 3-by-3 Hall sensor array. \textbf{(b)} Stereo and planar view of the soft tactile sensor, whose thickness is 5mm.\textbf{(c)} 3D histogram which reflexes the 9 raw values from the embedded 3-by-3 Hall sensor array. Each value represents the combination of the measured magnetic flux density from $x$, $y$ and $z$ axis.}
	\label{fig2}
\end{figure}

\begin{figure*}[htbp] \centering
	\begin{center}
		\includegraphics[width=\textwidth]{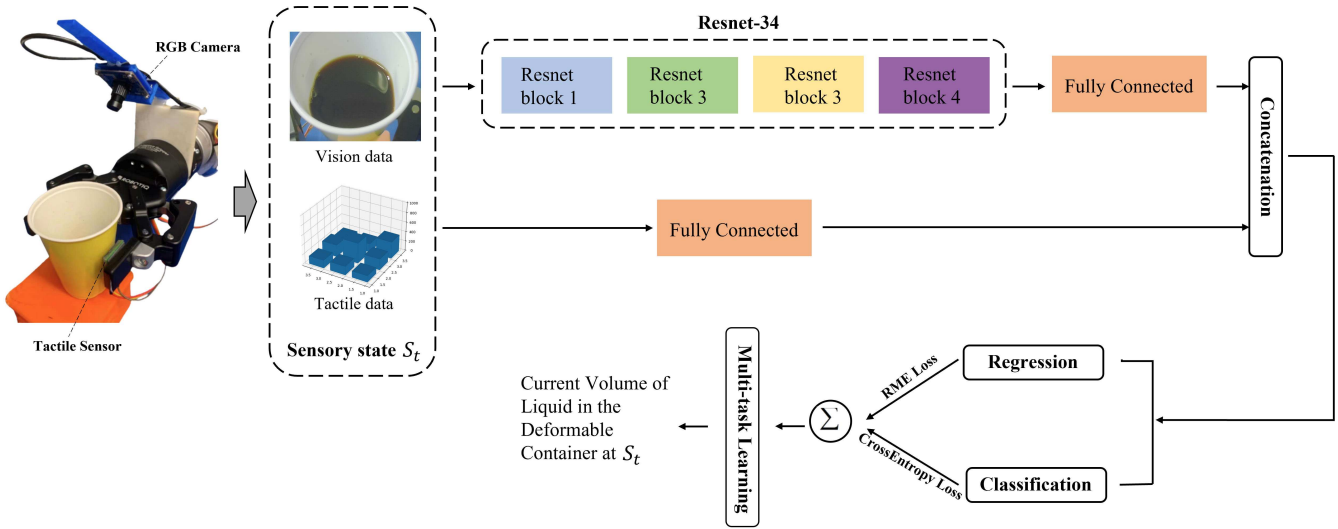}
	\end{center}
	\caption{ Network architecture of the deep visuo-tactile model. In the network, we combine the Cross-Entropy-Loss in classification with the MSE-loss in regression as the overall loss to do multi-task learning.}
	\label{fig1}
\end{figure*}

In our experiments we used a hardware configuration consisting of a 6-DOF collaborative UR5 arm, a Robotiq 2F-85 parallel gripper, a RGB web camera and a soft tactile sensor \cite{yan2021soft}, see the top-left of Figure \ref{fig1}. A RGB web camera was mounted above the gripper to provide visual cues about the real-time liquid level in the container. The soft tactile sensor consists of three layers (see Figure \ref{fig2}\textbf{(a)},\ref{fig2}\textbf{(b)}), which is motivated by the structure of human skin. The top layer is made of a flexible magnetic film. The middle layer is made of the soft silicone elastomer, which can sense the deformation of the magnetic film according to the change of magnetic flux densities. The bottom layer is a hard printed circuit board with a 3-by-3 Hall sensor array. We utilized the 27 raw magnetic flux densities for each set of the data (Each set of the data includes three magnetic flux densities ($B_{x}$,$B_{y}$ and $B_{z}$) of nine taxels.) and made the tactile sensor work at 10 Hz over an area of $18mm \times 18mm$ flexible magnet. When the liquid volume increases, the liquid level observed by the RGB camera and the magnetic flux densities measured by the tactile sensor \cite{yan2021soft} will both increase at the meantime. Based on the predictive model by fusing both vision and tactile cues, we explored the real-time liquid estimation when grasping a deformable container with liquid in a proprioceptive way, and expand our robotic system's functionalities to control and adjust the robotic grasp plan in real time according to the previous estimation.

\section{Deep Visuo-tactile Model for Real-time Liquid Estimation in Grasping}
\label{sec3:methodology}
By exploring the proprioception of the robot system, we formalize the real-time estimation of liquid in a deformable container as an end-to-end predictive model, in which the vision and tactile cues are fused. Based on the real-time estimation of liquid volume, the robotic system is automatically controlled and the grasping plan is adaptively adjusted in real time. When the container is stably lifted, the current volume of liquid in the deformed container will be estimated with the visuo-tactile predictive model. In the phase of evaluation, the outcome $o_{t}(V,L)$ are supposed to be estimated by the robot and compared to the target. At training time, as discussed in Section \ref{sec4:datacollection}, the ground truth volume $V_{gt}$ of liquid in the deformable container is automatically labelled as $L_{gt}$ by 10ml resolution for multi-task learning. The observation-outcome tuples($s_{i},o_{i}(V_{gt},L_{gt})$) are collected  to train an end-to-end model that can be used for the real-time liquid estimation when grasping a deformable container in the robotic system.

\subsection{End-to-end outcome prediction}
In a proprioceptive way, our aim is to obtain a function $f(\mathbf{s})$ which can predict the current volume of liquid in a deformable container grasped by the robotic system, given observation from the current grasp $\mathbf{s}$. The function $f$ is parameterized as a deep neural network as shown in Figure \ref{fig1} . For multi-modal inputs,  various design choices can be considered when designing the models \cite{ngiam2011multimodal}. In our experiments, we designed a network to process the state $\mathbf{s}$, which consists of the raw visual inputs from the mounted RGB web camera in four deep stacks of convolutional layers and one fully-connected layer. Additionally, the magnetic flux densities from the soft tactile sensor \cite{yan2021soft} is processed in a fully-connected layer. As shown in Figure \ref{fig1}, we subsequently fuse cues from different modalities as follows: the  vectors of image feature and tactile values are concatenated as one vector, which is simultaneously fed to two different tasks: classification and regression, and obtain two different losses. We then combine the losses with weights $\lambda_{1}$ and $\lambda_{2}$ as an overall loss to do multi-task learning and produce the estimated volume $f(\mathbf{s_{t}})$ of the liquid at $\mathbf{s_{t}}$. With the collected data $X$, we aim to train a network $f$  in which the loss $L(f,X)=\sum_{(\mathbf{s},o)\in X} l_{overall}(f(\mathbf{s},o))$ is minimized. Here, $l_{overall}$ is the combination of weighted mean-squared-error loss and cross-entropy loss. 

\begin{itemize}
	\item [\textit{a)}] \textit{Design of the network:} Each RGB image is processed by the proposed convolutional network. It is worth noting that we utilize the penultimate layer of ResNet-34 (\cite{he2016deep}, a 34-layer deep residual network) and further separately pass the features produced by ResNet-34 \cite{he2016deep} and the raw magnetic flux densities to a fully-connected (FC) layer with 512 and 27 hidden units. To fuse these features, we concatenate the FC layers from two branches (vision and tactile data), and then simultaneously pass them through a pre-defined classification and regression network. Finally, we combine the weighted mean-squared-error(MSE) and cross-entropy losses produced by last step to do multi-task learning that estimate current volume of the liquid in the deformable container. The proposed architecture of model in our system is shown in Figure \ref{fig1}.
	
	\item [\textit{b)}] \textit{Training parameters:} We pre-train the network by deploying the weights from a object classification model trained on ImageNet \cite{deng2009imagenet} to speed up training. We subsequently perform the model optimization with a batch size of 32 and the training epoch is 100 (training on a dataset of 2581 examples). The start learning rate $lr_{start}$ in our experiments is 0.001 and we used the MultiStepLR scheduler to adjust the learning rate in training after 40 epochs and 70 epochs, separately lowering the learning rate with a factor $\gamma =0.1$.
	
\end{itemize}

\subsection{Multi-task learning}
We have a multi-modal robotic system which consists of visual and tactile cues. In our model, we apply the multi-task learning techniques by combining the loss in classification and regression.

In classification, we took Cross-Entropy-Loss as follows:
\begin{equation}
\centering
\begin{aligned}
l_{CrossEntropy}(x,class)
=-log(\frac{exp(x[class])}{\sum{exp(x[j])}}) \\
=-x[class]+log(\sum{exp(x[j])})
\end{aligned}
\label{eq6.1}
\end{equation}
as the criterion for model optimization. \textbf{$x$} represents the estimated label and \textbf{$class$} represents the ground truth label.

In regression, we apply MSE-Loss (Mean-Squared-Error loss):
\begin{equation}
\centering
\begin{aligned}
l_{MSE}(x,y)=mean(L)\\
L=\{l_{1},...,l_{N}\}^\top, l_{n}=(x_{n}-y_{n})^2,
\end{aligned}
\label{eq6.2}
\end{equation}
as the criterion into the optimization of regression. $x$, $y$, $N$ represent the input, target volume and the batch size(we set $N=32$), respectively. 

To apply multi-learning techniques, we combine the Cross-Entropy-Loss $l_{CrossEntropy}$ in classification with the MSE-Loss $l_{MSE}$ in regression as the overall loss $l_{overall}$:
\begin{equation}
\centering
\begin{aligned}
l_{overall}= \lambda_{1} l_{CrossEntropy} + \lambda_{2}l_{MSE}
\end{aligned}
\label{eq6.3}
\end{equation}

\subsection{Robot proprioceptive capability in liquid estimation and control}
Based on the robot's proprioceptive capability in real-time liquid volume estimation, we can serve the expected volume $V^{*}$ of liquid in the deformable. When the water pump is working, we can estimate the current volume $V_{t}$ of the liquid at sensory state $\mathbf{s_{t}}$, and control the water pump's working state $P_{t+1}$  in sensory state $\mathrm{\mathbf{s_{t+1}}}$ as:

\begin{equation}   P_{t+1} =
\begin{cases}
1 ,  &  \text{if $V^{*} - V_{t}$ \textgreater  $0$,} \\
0 ,  &  \text{if $V^{*} - V_{t} \leq 0$}
\end{cases}  
\label{eq4.1}
\end{equation}

where 0 represents 'off' and 1 represents 'on' for the water pump. Once the volume of liquid has satisfied the expectation, the water pump will be automatically stopped by the robotic system.

\subsection{Real-time grasping plan adjustment}
We use the visuo-tactile model $f$ to adjust the grasping force in real time to minimize the deformation of  container and guarantee the stability of grasping, moving and possible manipulation. For example, when the liquid is continuously flowing into a deformable container, the weight is changing. If the grasping force is not further tuned, the system may not be stable and the container may get out of control. Liquid inside will be spilt out under this circumstance. We propose a strategy
for grasping plan adjustment according to the estimation of current volume:
\begin{equation}
\centering
\mathbf{a_{t}}=\Delta( \pi_{f(s_{t})}, \pi_{0} )
\label{eq4.2}
\end{equation}

$\pi_{0}$ is the current grasping plan, which remains unchanged since grasping initialization. $\pi_{f(s_{t})}$ is supposed grasping plan, which is discretely mapped from the estimated volume of liquid in the deformable container in real time. $\mathbf{a_{t}}$ is the robot action from current to the supposed grasping plan.

\section{Data Collection and Processing}
\label{sec4:datacollection}
\subsection{Data collection}
To collect the data necessary to train our model, we designed an automated data collection process. In each trial, RGB images from a Webcam mounted on the robotic arm was used to approximately estimate the position of the deformable container and the liquid inside it. Then we set the coordinates ($x,y$) of the gripper to the position of the container and set the height of the gripper to a certain value between the height of the container and the floor in each trial. After moving to the estimated position, the gripper closes with a certain gripping force $F$ and attempt to grasp the container. The container is then lifted to a specific height and the end-effector maintains the same configuration for 5 seconds before further manipulations. Meanwhile, both the visual cues from the webcam and the magnetic flux density data from the tactile sensor \cite{yan2021soft}, which is attached to one fingertip of the gripper, are collected.
Based on the pre-measured volume of the liquid in the container, the amount of liquid inside the container was automatically labelled with 10 ml resolution (i.e., 0-10ml was labelled as 0, 11-20ml was labelled as 1 etc.). Due to the material property of the magnetic film on the tactile sensor, there is enough friction between the tactile sensor and the container to balance the gravity. Despite the volumes of the liquid vary in different trials,  occasional slips rarely occurred in the process of data collection. So the training data are generally collected in the stable grasps. At a same timestamp, a RGB image and the raw magnetic flux density data, which contains 27 values, are aligned and collected as a pair. Since the tactile sensor\cite{yan2021soft} is 3*3 grid-structure and contains 9 embedded Hall sensors. To guarantee the generalization of our model, we add a small perturbation in grasp trials to make the container contact with different areas of the tactile sensor. Consequently, we collected 110 grasping trials in total over the same container with different volumes of liquid. Our dataset contains 2581 examples.

\subsection{Data processing}
We first re-scale the raw RGB images as $256 \times 256$. For the purpose of data augmentation, subsequently, we perform the $224 \times 224$ random crops sampling on the visual cues. Although the resolution will be substantially lower than the original one of the web camera, it is a standard image resolution for classification with ResNet-based model in computer vision. In this work, although we did not discuss how the image resolution will affect the systematic performance, it is an interesting topic in future. In the phases of both data collection and experimental evaluation, the raw visual and tactile data are normalized. In our experiments we noticed that the initial values of the tactile sensor may vary in different grasp trials. However, after conducting efficient different trials in the phase of data collection, the initial values of the sensor did not seem to exercise a great influence over the performance any more. It indicates that the features learned by the model are independent of the initial values of the tactile sensor.

\section{Experimental Validation}
\label{sec5:experimentalValidation}

To validate our estimation model in the robotic grasping system with multiple sensing modalities, we first perform the evaluation of our model with the collected dataset. Then we compare the model in a real robot grasping scenario, and test its generalization capabilities when the liquid is continuously added into the container and the volume of the liquid is estimated in real time. Moreover, we present the robotic grasping and manipulation in the scenario of a service robot and complete the task of filling the deformable container with a specific volume of liquid based on our visuo-tactile model. Finally, we demonstrate that it is possible to correspondingly adjust the applied force in real time to decrease the deformation of the container while maintaining a stable grasp. To show the performance of robotic grasping and manipulation, we prepare the online video demos at: \href{https://youtu.be/UbvK3O4ypHs}{https://youtu.be/UbvK3O4ypHs}

\subsection{Model evaluation}
We have a multi-modal robotic system which consists of visual and tactile cues. Each raw visuo-tactile observation $s$ is acquired from the soft tactile sensor \cite{yan2021soft} and the mounted RGB camera, as shown in Figure \ref{fig1}. In the initialization, the gripper grasps the deformable container with a specific force $F$. Due to the material property of the magnetic film on the tactile sensor, the friction between the tactile sensor and the container is almost enough to balance the gravity and occasional slips rarely occurred in the initialization phase. Hence we did not discuss the occasional slips here. We separately evaluate the performance of classification, regression and multi-task learning.

\textbf{Classification.} First, we seek to separately evaluate the performance of volume classification (classified by each 10 ml) with  vision, tactile and the visuo-tactile (fusion of the vision and tactile data) inputs. As mentioned in Section \ref{sec4:datacollection}, the volume data have been automatically labeled with 10ml resolution based on the pre-measured ground truth. We apply the ResNet-34 into our classification model. The start learning rate $lr_{start}$ in our experiments is 0.001 and we used the MultiStepLR scheduler to lower the learning rate in training separately after 40 epoches and 70 epoches with a factor $\gamma =0.1$. The optimizer we used in training is SGD and batch-size we utilized for optimizing the model is 32. Following Equation \ref{eq6.1}, we separately evaluated the performance of different variations (vision-only, tactile-only and vision + tactile) for our classification model using the labeled dataset. The result of K-fold($K$=3) cross-validation is reported in the $1^{st}$ row of Table \ref{table1}. 

\textbf{Regression.} Similarly, we then evaluate our regression models of liquid volume estimation separately trained with vision, tactile and the fusion of above two. The learning parameter values, including learning rate, scheduler and optimizer, are exactly the same as the ones used in the classification model. However, we utilized the exact normalized volumes as the ground truth in training instead of labels. With Equation \ref{eq6.2}, the results of K-fold cross-validation of regression model are reported in the $2^{nd}$ row of Table \ref{table1}.

\textbf{Multi-task Learning.} Last but not least, we evaluated the performance of multi-task learning techniques by Equation \ref{eq6.3}. In our experiments, because the cross-entropy-loss  $l_{CrossEntropy}$ is much greater than the MSE-loss $l_{MSE}$, we rescale them in the overall loss and set the parameter $\lambda_{1}$ and $\lambda_{2}$ in Equation \ref{eq6.3} as 1 and 100, respectively. With multi-task learning techniques, we separately trained the models with vision, tactile and visuo-tactile data. The results of K-fold cross-validation of multi-task learning model are reported in the $3^{rd}$ row of Table \ref{table1}.

\begin{table}[ht]
	\caption{K-fold (K=3) cross-validation errors ({\upshape mean} $\pm$ {\upshape std. err.})  of volume ({\upshape ml}) estimation for the different models trained with 2581 data points.}
	\centering 
	\scalebox{0.9}{
		\begin{tabular}{c c c c}
			\hline\hline                        
			& Vision  & Tactile  & Vision+Tactile \\ [0.5ex]
			\hline                  
			Classification & 7.460 $\pm 0.030$ & 8.350 $\pm 0.045$ & 6.025 $\pm 0.025$ \\
			Regression & 3.874 $\pm 0.019$ & 10.119 $\pm 0.015$  & 2.160 $\pm 0.018$  \\
			Classification+Regression & 3.475 $\pm 0.023$& 4.715 $\pm 0.019$  & 1.972 $\pm 0.014$  \\ [1ex]      
			\hline
		\end{tabular}
	}
	\label{table1}
\end{table}

To summarize, we see that errors indeed drop significantly when the vision and tactile cues are fused, validating that the visuo-tactile model can successfully learn from both visual and tactile information and is effective enough to improve the estimation performance. From another perspective, to compare different learning techniques, including classification, regression and multi-task (classification + regression), the model trained with multi-task learning techniques obviously outperform others. Finally, with the fusion of vision and tactile data, we obtain the best model by applying multi-task learning.

\subsection{Evaluation of robot proprioceptive capability in liquid estimation and control}
\label{subsection:proprioception}
\begin{figure}[htbp] \centering
	\begin{center}
		\includegraphics[width=0.33\textwidth]{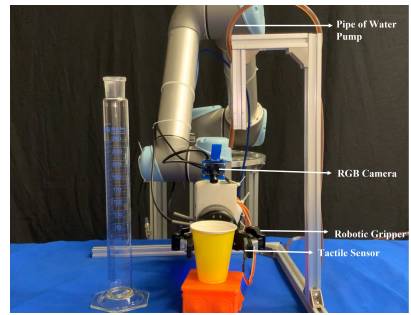}
	\end{center}
	\caption{The experimental setup for evaluating robotic proprioceptive capability in liquid estimation and control when filling the expected volume of liquid into a deformable plastic container. A soft tube connected with a water pump continuously provides liquid flow until the estimated liquid volume in the deformable container equals to the expected one.}
	\label{experiments}
\end{figure}

Next, we evaluate the predictive models on the real robotic tasks. In the experimental evaluations, we had the robot grasp and move the deformable container to a designated position to fill the container with a specific volume of liquid. The experimental setup is shown in Figure \ref{experiments}. First, the robotic gripper approaches to and grasps the deformable container with a determined grasping plan. Then the container is lifted and liquid starts to flow into the container. There is a soft tube, which is connected with a controllable water pump, continuously providing the liquid flow. When the liquid flows into the container, simultaneously the current liquid volume is constantly estimated in real time with robot's proprioception from the learned visuo-tactile model. In Equation \ref{eq4.1}, the water pump's working state $P_{t+1}$ in sensory state $s_{t+1}$ is determined by the current volume of liquid in $s_{t}$. The water pump we used has been well calibrated and its working state can be switched instantly by the robotic system. Hence the systematic error caused by the delay of water pump will not be discussed here. Once the estimated volume of liquid reach the expected one, the water pump is suspended by the robotic system.

\begin{figure*} 
	\centering 
	\subfigure[Expected volume is 40 ml]{ 
		\label{result-40}
		\includegraphics[width=2.24 in]{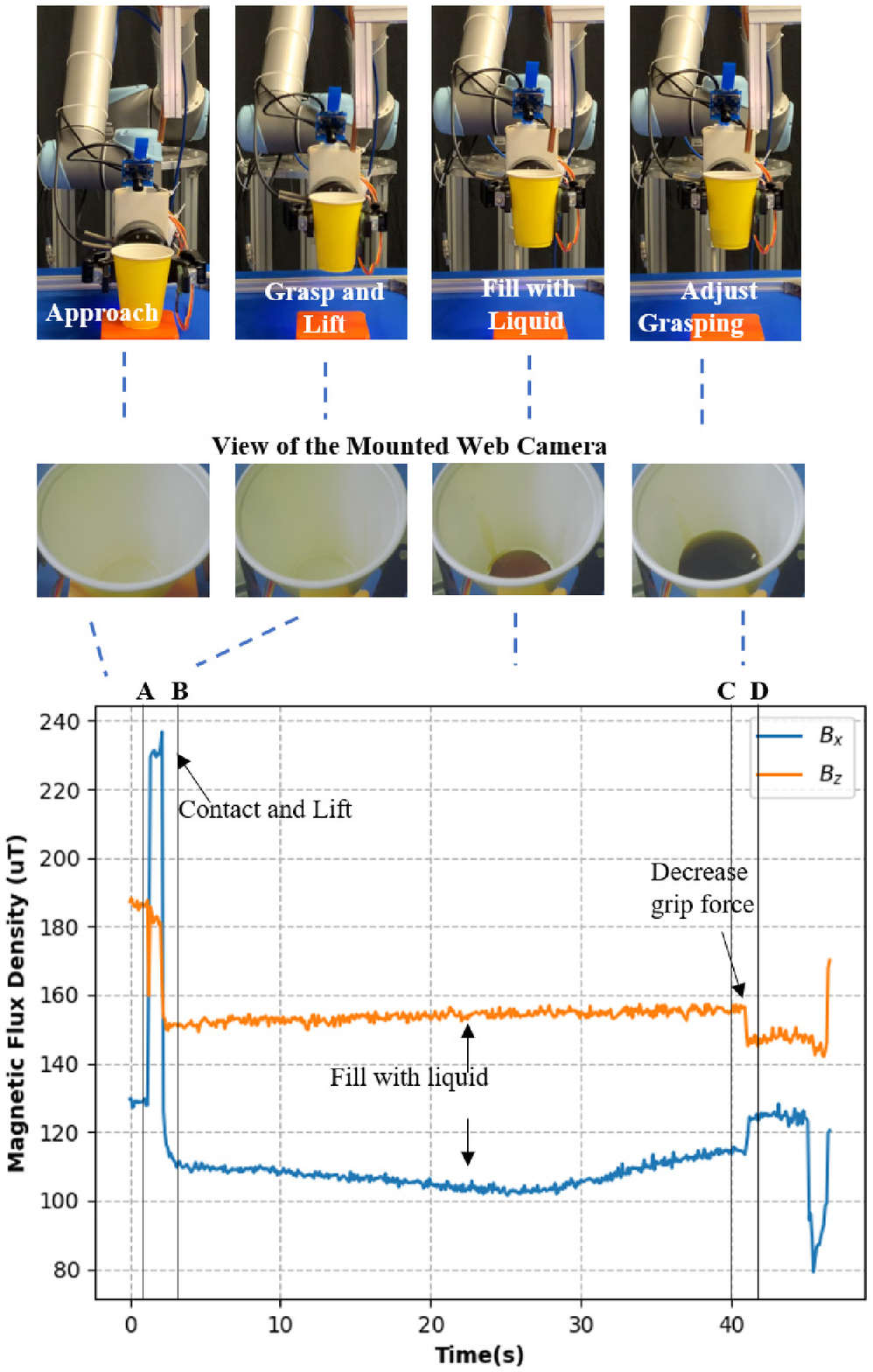} 
	} 
	\subfigure[Expected volume is 80 ml]{ 
		\label{result-80}
		\includegraphics[width=2.24 in]{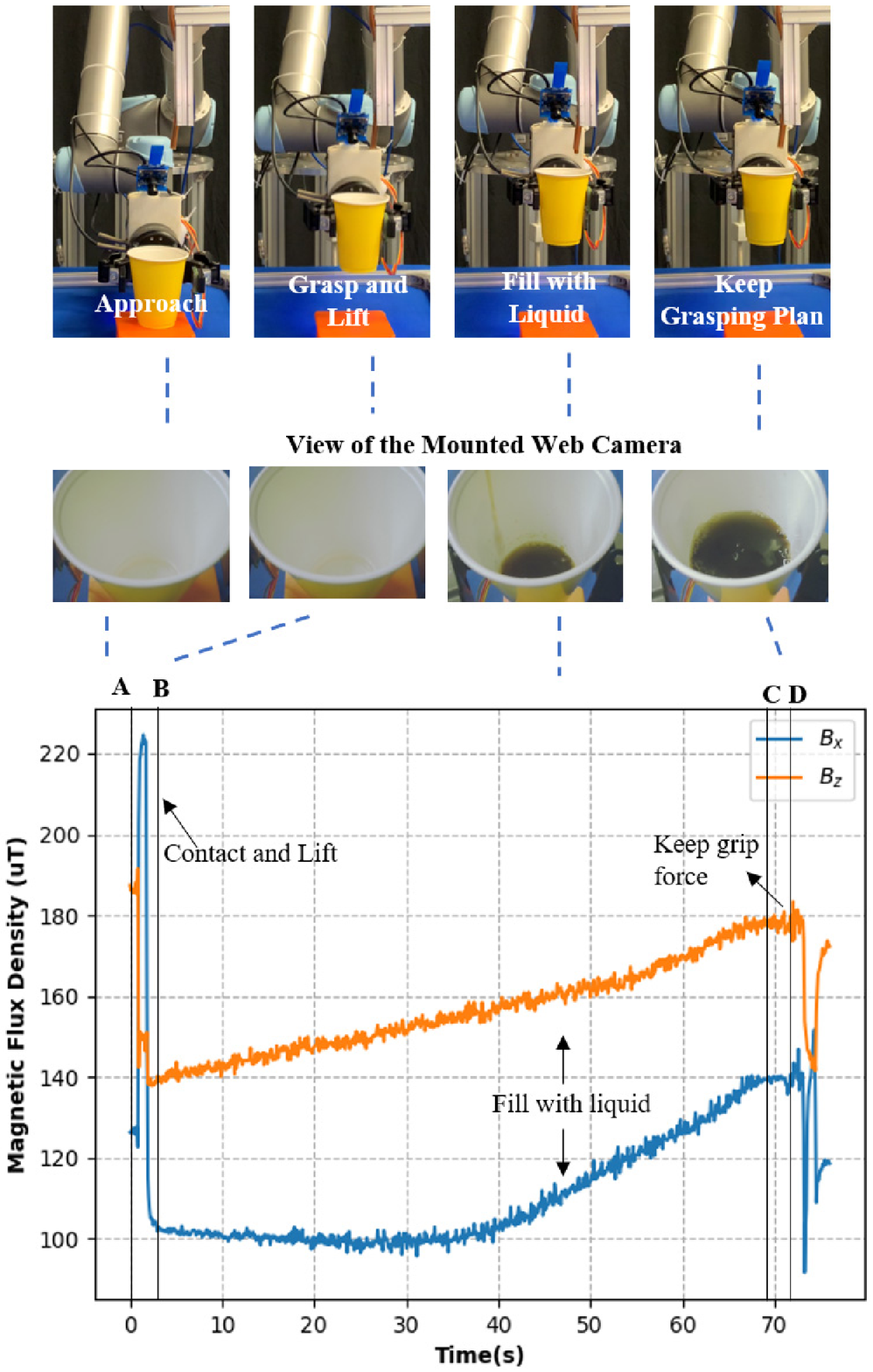} 
	} 
	\\
	\subfigure[Expected volume is 120 ml]{ 
		\label{result-120}
		\includegraphics[width=2.24 in]{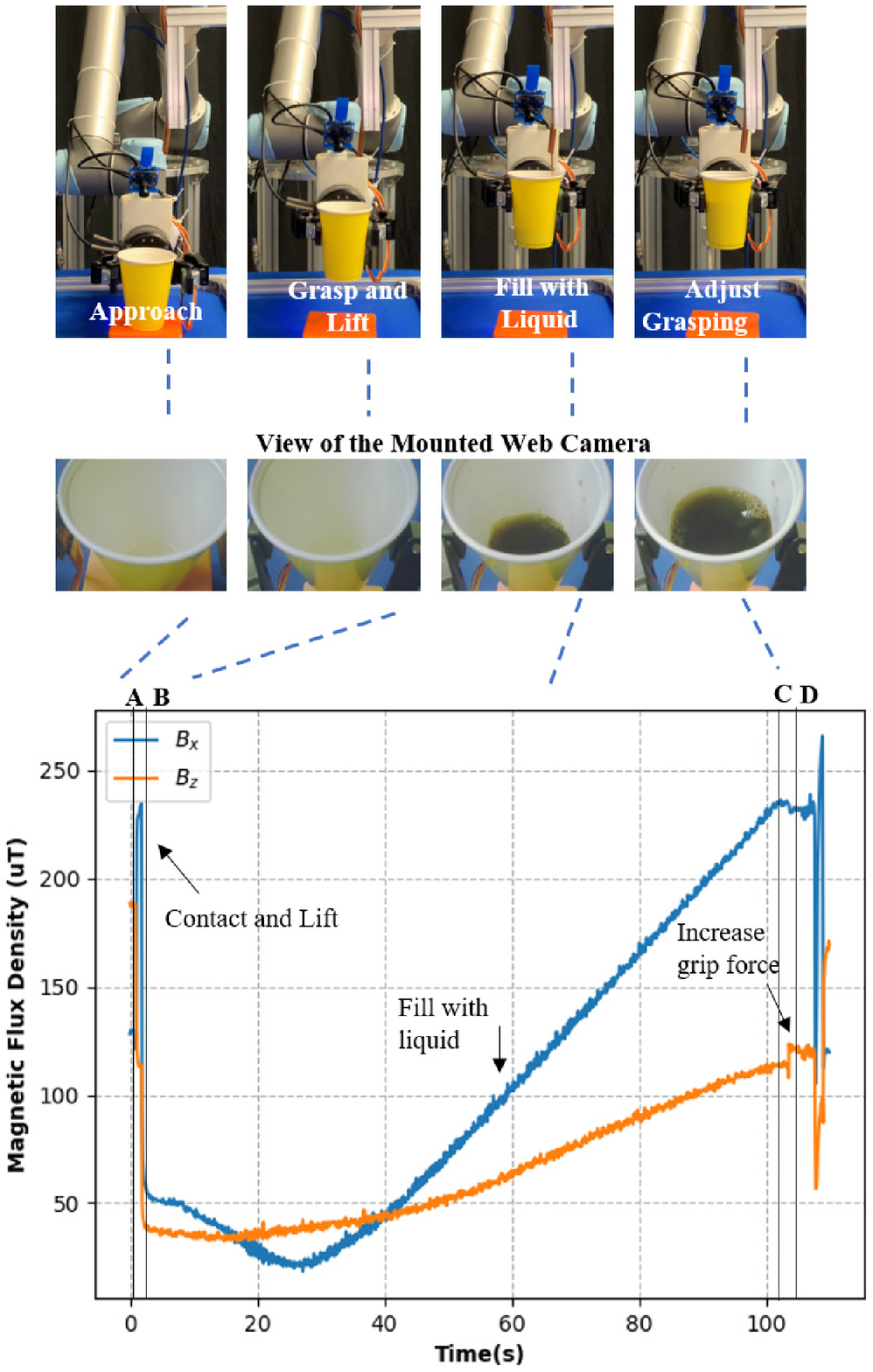} 
	} 
	\subfigure[Expected volume is 140 ml]{ 
		\label{result-140}
		\includegraphics[width=2.24 in]{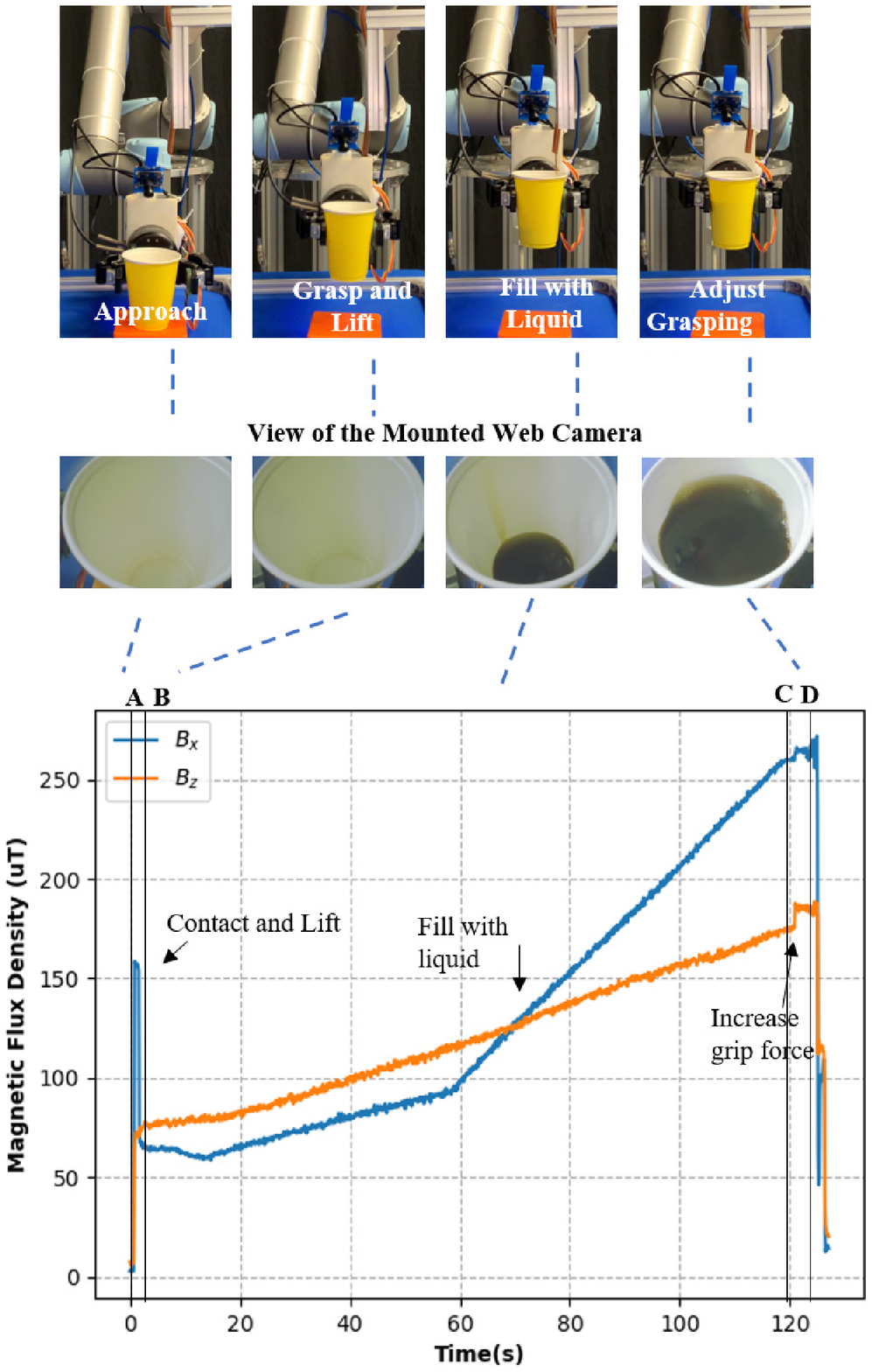} 
	}
	\caption{Experimental validations of the real-time liquid volume estimation. In each sub-figure, the state of the robot, the view of mounted RGB camera and the magnetic flux densities $B_{x}$ (along the gravity) and $B_{z}$(perpendicular to the sensor surface) are separately shown in each row. \textbf{(a)} The expected volume of liquid is 40 ml. The liquid actually filled is 41 ml. \textbf{(b)} The expected volume of liquid is 80 ml. The liquid actually filled is 80 ml. \textbf{(c)}The expected volume of liquid is 120 ml. The liquid actually filled is 122 ml. \textbf{(d)} The expected volume of liquid is 140 ml. The liquid actually filled is 142 ml. For more experimental details, please refer to the online video at: \href{https://youtu.be/UbvK3O4ypHs}{https://youtu.be/UbvK3O4ypHs}} 
	\label{results}
\end{figure*}

\begin{figure*}[htbp] 
	\centering
	\begin{center}
		\includegraphics[width=0.9\textwidth]{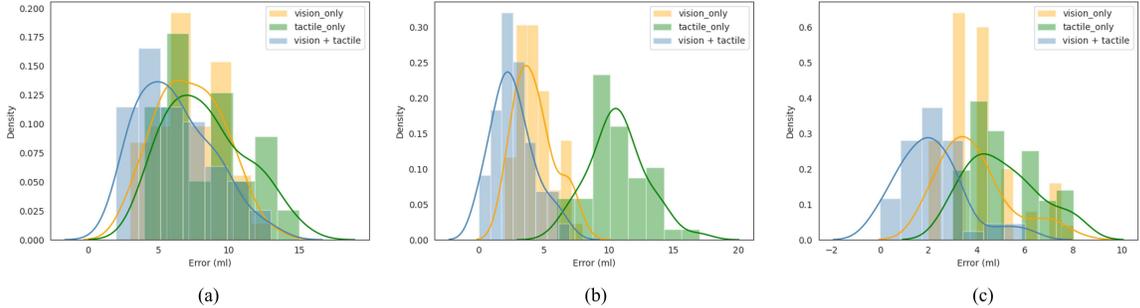}
	\end{center}	
	\caption{Histogram and density plot for the measured errors of different models and variants of each model. Vision + tactile variant performs best in each model. \textbf{(a)} In classification model, the mean errors of vision-only, tactile only and tactile + vision variants are 7.24 ml, 8.34 ml, 6.18 ml. \textbf{(b)} In regression model, the mean errors of vision-only, tactile only and tactile + vision variants are 4.28 ml, 10.68 ml, 2.66 ml. \textbf{(c)} In classification + regression model, the mean errors of vision-only, tactile only and tactile + vision variants are 3.9 ml, 5.08 ml, 1.98 ml.}
	\label{histogram}
\end{figure*}

To validate the performance of robot's proprioception from the learned visuo-tactile model, we try to fill the deformable container with different expected volumes and validate the accuracy with the graduated cylinder in Figure \ref{experiments} after the water pump is suspended. Figure \ref{results} shows 4 different experimental tests. The RGB image from the mounted web camera and the raw magnetic flux densities measured by the tactile sensor dynamically change as the liquid volume increases in real time. When the gripper successfully contacted and lifted the container, in each sub-figure of Figure \ref{results}, spike appears in the magnetic flux densities $B_{x}$ (along the direction of gravity). After the success of lifting is detected, with robot's proprioceptive capability, the estimation of liquid volume and water pump working control simultaneously start in the robotic system. As Figure \ref{results} shows, in each test, the magnetic flux densities incrementally change as the liquid flows into the container, until the volume reaches the expectation and the water pump is automatically suspended by the robotic system. In each test, we initially apply the same grasping plan, including grip force and the contact position on the container. We recorded the ground truth of liquid volume $V_{g}^{i}$ after each test and statistically analyzed the error $E_{i}$ between the expected volume $V_{e}^{i}$ and the ground truth $V_{g}^{i}$:

\begin{equation}
\centering
\begin{aligned}
E_{i}=\left|V_{e}^{i} - V_{g}^{i}\right|
\end{aligned}
\label{eq6.4}
\end{equation}

We have evaluated 3 different models on the robotic system: classification, regression and multi-task learning (regression + classification). Moreover, we compare the fusion of visual and tactile data with the vision-only and tactile-only variants of each model (see Figure \ref{histogram}). The mean values of each model for different variants are statistically analysed in the Table \ref{table2} by separately computing the mean value of $E_{i}$ in $N$ tests ($N$=50). To summarize, the fusion of vision and tactile with the multi-task learning model (classification + regression) shows the best performance.

\begin{table}[ht]
	\caption{Average evaluated errors(ml) in $N$ tests for different variants of different models on the robotic system ($N$=50 for each evaluation).}
	\centering
	\scalebox{0.9}{
		\begin{tabular}{c c c c}
			\hline\hline                      
			& Vision  & Tactile  & Vision+Tactile \\ [0.5ex]
			\hline                  
			Classification & 7.24  & 8.34  & 6.18 \\
			Regression & 4.28  & 10.68 & 2.66   \\
			Classification+Regression & 3.9 & 5.08   & 1.98   \\ [1ex]      
			\hline
		\end{tabular}
	}
	\label{table2}
\end{table}

\begin{figure}[htbp] \centering
	\begin{center}
		\includegraphics[width=0.4\textwidth]{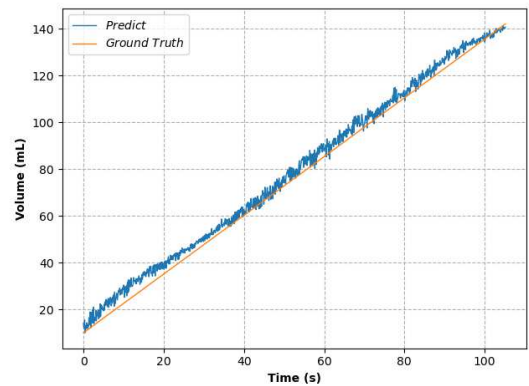}
	\end{center}
	\caption{The result of real-time liquid volume estimation when utilize the visuo-tactile model with multi-task learning techniques. We take $V_e$ = 140 ml (expected liquid volume) as an example. The orange line represents the computed ground truth (supposing the water pump works at a constant speed). The blue line represents the result of real-time liquid volume estimation. }
	\label{estimation}
\end{figure}

We explicitly visualize the real-time liquid estimation result when filling the container with expected volume $V_{e}$ of liquid (we take $V_e$ = 140 ml as an example), see Figure \ref{estimation}. Supposing the water pump works at a constant speed, we obtained the ground truth of liquid volume. Meanwhile, since the liquid volume that we focused on and trained the network for is above 10, we plot the volume estimation result from 10 in Figure \ref{estimation}. At the end of water-filling task, the volume of liquid actually filled is 142 ml.

\subsection{Real-time grasping plan adjustment  }
As we mentioned in Section \ref{subsection:proprioception}, we initially apply the same grasping plan $\pi_{0}$ to grasp the container. After the container is filled with a certain volume of liquid, the initial grasping plan becomes insufficient. When the liquid volume is small but large grip force is applied, the container will be excessively deformed. On the contrary, if current grip force is not sufficient for successfully manipulate the deformable container with a large volume of liquid, the container may slip or rotate. Such uncontrollable conditions always result in failed manipulation. An adaptive real-time grasping plan should be considered.

According to Equation \ref{eq4.2}, the strategy for grasping plan adjustment depends on the current estimation of liquid volume. We generally define two threshold $\theta_{1}$ and $\theta_{2}$. If the estimated liquid volume $f(s_{t})$ is less than $\theta_{1}$, the grip force will be decreased. If $f(s_{t})$ is more than $\theta_{2}$, the grip force will be increased. Otherwise, the grip force will remain the same. In our experiment, $\theta_{1}=50$ and $\theta_{2}=100$. In Figure \ref{results}, the real-time grasping plan adjustment is explicitly reflected by the magnetic flux densities $B_{z}$, which is perpendicular to the sensor surface. In our experiments, after the adjustment of grasping plan, the success rate of subsequent manipulation (rotation, translation ,etc.) has increased by 8\%.

\section{Conclusion}
\label{sec6:conclusion}
Touch sensing is an inherently active sensing modality which can assist the robot to sensitively feel the changes in the environment with feedback controller that incorporates tactile inputs in the process of grasping. It is of great challenge to design the controller in such an active fashion. especially when the tactile sensing modality is combined with visual inputs. In this paper, we proposed an end-to-end approach for predicting current state of liquid in a deformable container using raw visual and tactile inputs with a mounted RGB camera, which provides visual cues and a tactile sensor \cite{yan2021soft}, which provides raw magnetic flux densities. Instead of utilizing auxiliary devices, when the liquid source is unknown, the visuo-tactile model possesses the robot of proprioception to estimate the volume of liquid in real time . To train the predictive model, we performed data collection from 110 grasping trials in total over the same container with different volumes of liquid. The learned model is capable of estimating current volume of liquid while the liquid is continuously flows into the deformable container. The results indicate that our visuo-tactile model substantially improves the accuracy of real-time liquid estimation compared to models that are based on only a single sensing modality (e.g., vision or tactile sensing) or a single technique (e.g., classification or regression). To further validate the result, we perform a real-world evaluation of different models in active liquid-filling tests. The average estimation error in our tests in around 2 ml, which is acceptable and obviously outperform other models. Furthermore, we demonstrated that with the proposed visuo-tactile model, it is feasible to adjust the grasping plan in real time by adaptively decreasing grip force while preserving the success of grasping and increasing grip force to increase the success rate of subsequent manipulation by 8 \%.

Our method has some limitations that could be addressed in future work. First, our visuo-tactile model only performs single-step predictions, instead of utilizing temporally gathered information. Second, our current model does not explicitly consider the reaction to slipping during the lift-off, hence not taking advantages of the interactive nature of tactile cues in grasping. Third, the performance of our system may decrease when the liquid rushes at a high rate. As future work, we would like to explore solutions to the information-gathering model, more interactive reactions in grasping and more stable and accurate estimation approach even when the flow rate is high.

\bibliographystyle{./IEEEtran}
\bibliography{./IEEE}

\end{document}